\documentclass[10pt,twocolumn,letterpaper]{article}

\usepackage{cvpr}              % To produce the CAMERA-READY version
\usepackage{bm}
\usepackage{tikz}
\usepackage{xcolor}
\usepackage{graphicx}

\usepackage{pgfplots} 
\usepackage{pgfplotstable}
\pgfplotsset{compat=1.16}
\usepackage{caption}

\usepackage{listings} % for pytorch code
\usepackage{xcolor} % to define our colors
\usepackage{lineno}
\usepackage{pifont} % to use \xmark and \checkmark commands
\usepackage{multirow} % used for multiple column table

\definecolor{cvprblue}{rgb}{0.21,0.49,0.74}
\usepackage[pagebackref,breaklinks,colorlinks,allcolors=cvprblue]{hyperref}

\def\paperID{*****} % *** Enter the Paper ID here
\def\confName{CVPR}
\def\confYear{2025}

\title{DyCON: Dynamic Uncertainty-aware Consistency and Contrastive Learning for Semi-supervised Medical Image Segmentation}

\author{Maregu Assefa, Muzammal Naseer, Iyyakutti Iyappan Ganapathi, Syed Sadaf Ali \\ Mohamed L Seghier, Naoufel Werghi \\
C2PS - Khalifa University of Science and Technology, Abu Dhabi, UAE
}

\begin{document}
\maketitle
\begin{abstract}
Semi-supervised learning in medical image segmentation leverages unlabeled data to reduce annotation burdens through consistency learning. However, current methods struggle with class imbalance and high uncertainty from pathology variations, leading to inaccurate segmentation in 3D medical images. To address these challenges, we present DyCON, a Dynamic Uncertainty-aware Consistency and Contrastive Learning framework that enhances the generalization of consistency methods with two complementary losses: Uncertainty-aware Consistency Loss (UnCL) and Focal Entropy-aware Contrastive Loss (FeCL). UnCL enforces global consistency by dynamically weighting the contribution of each voxel to the consistency loss based on its uncertainty, preserving high-uncertainty regions instead of filtering them out. Initially, UnCL prioritizes learning from uncertain voxels with lower penalties, encouraging the model to explore challenging regions. As training progress, the penalty shift towards confident voxels to refine predictions and ensure global consistency. Meanwhile, FeCL enhances local feature discrimination in imbalanced regions by introducing dual focal mechanisms and adaptive confidence adjustments into the contrastive principle. These mechanisms jointly prioritizes hard positives and negatives while focusing on uncertain sample pairs, effectively capturing subtle lesion variations under class imbalance. Extensive evaluations on four diverse medical image segmentation datasets (ISLES'22, BraTS'19, LA, Pancreas) show DyCON's superior performance against SOTA methods\footnote{\href{https://dycon25.github.io/}{\textcolor{red}{Project Page: \url{https://dycon25.github.io/}}}}.
\vspace{-1em}
\end{abstract}
    
\section{Introduction}
\label{sec:intro}
Medical image segmentation plays a critical role in the diagnosis and timely treatment of neurological diseases, especially in complex and irregular regions like brain lesions. However, manual segmentation is labor-intensive and error-prone due to the intricate brain structures and the variable nature of lesions. Deep learning models~\cite{UNet,UNETR,SwinUNETR,Factorizer} offer a promising solution, however, their reliance on large-scale annotated datasets is a major obstacle. Semi-supervised learning (SSL) addresses this by utilizing unlabeled data with minimal labels, primarily through consistency regularization~\cite{ua_mt,mean_teacher,xu:ambiguity,xu:cross_modal_consistency} and pseudo-labeling~\cite{self_loop_pseudo_label,bai:pseudo_labeling,wu:semi_pseudo_label,pseudo_label_threshold}.
\begin{figure}[!t]
    \centering
    \begin{tikzpicture}
        \begin{axis}[
            ybar=0pt,
            bar width=0.65cm,
            width=7cm,
            height=5cm,
            ymin=0,
            ymax=90,
            ylabel={Performance},
            symbolic x coords={MT, CT, MT+DyCON, {CT+DyCON}},
            xtick=data,
            x=1.65cm, % Increases space between x-axis labels
            nodes near coords,
            nodes near coords style={font=\small},
            enlarge x limits=0.2,
            legend style={at={(0.26,0.84)}, anchor=south, legend columns=-1},
            ylabel style={yshift=-0.5em},
            axis y line*=left,
            axis x line*=bottom,
            ymajorgrids=true,
            grid style=dashed,
            xticklabel style={font=\footnotesize}, % Smaller font for x-axis labels
            every node near coord/.append style={font=\footnotesize}
        ]
        % Dice scores (light skyblue)
        \addplot[draw=none, fill={rgb,255:red,135; green,206; blue,250}] coordinates {
            (MT, 59.10) (CT, 63.32) (MT+DyCON, 76.38) (CT+DyCON, 73.73)
        };
        % HD95 scores (light salmon)
        \addplot[draw=none, fill={rgb,255:red,250; green,128; blue,114}] coordinates {
            (MT, 22.01) (CT, 27.70) (MT+DyCON, 10.38) (CT+DyCON, 19.41)
        };
        \legend{Dice(\%)$\uparrow$, HD95$\downarrow$}
        \end{axis}
    \end{tikzpicture}
    \vspace{-1em}
    \caption{Comparative performance (average Dice (\%) and HD95 on ISLES'22 and BraTS'19) when integrating DyCON into MT (Mean-Teacher) and CT (Co-Training) frameworks with 10\% labeled data. DyCON improves segmentation accuracy without modifying training pipelines, proving its versatile effectiveness.
    }
    \label{fig:dycon_performance}
    \vspace{-1.5em}
\end{figure}
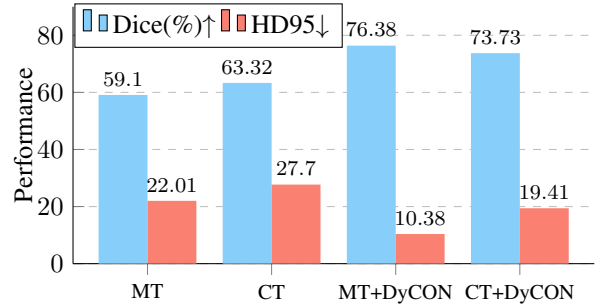
\vspace{-0.2em}
However, mainstream SSL methods face inherent challenges in radiology image segmentation, such as extreme class imbalance, where lesion voxels are significantly less than background voxels, and high uncertainty due to pathology variations, including differences in lesion size and shape. These factors result in inaccurate and inconsistent segmentation with poorly defined lesion boundaries. In response, recent methods incorporate uncertainty estimation techniques to guide the consistency learning~\cite{upc,fussnet_uncertainty,ct_lesion_uncertainty,self_loop_pseudo_label,consistency_dyn_threshold,ua_mt}. They typically rely on Monte Carlo dropout or deep ensembles to produce multiple predictions and filter out unreliable ones based on thresholds or variance metrics. Some approaches~\cite{wang:mcf,xu:dual_uncertainty_guided,zhu:hybrid_double_uncertainty,huang:dual_local_contrastive_consistency} further use dual-decoders or sub-networks to generate discrepancy masks to guide the consistency loss or displace uncertain regions~\cite{chi:ABD}.

Despite promising results, these methods inherently discard voxels with high uncertainty due to filtering or dual-decoder/sub-network disagreements. By limiting the consistency loss to certain localized regions, these methods fail to leverage the global imaging context. Consequently, resulting in suboptimal segmentation in cases with scattered or irregular lesion distributions (shown in Fig.~\ref{fig:scattered_voxels} in Section~\ref{subsec:ablation}). Additionally, contrastive learning has gained increasing prominence in medical image segmentation~\cite{liu:prototype_pseudo_labeling,huang:dual_local_contrastive_consistency,chaitanya:local_contrastive,tang:semi_hard_pos_contrast}, showing strong potential to improve feature discrimination. However, these contrastive methods face critical challenges, including sampling bias~\cite{chuang:debiased_contrastive} and class collision~\cite{chaitanya:contrastive_global_local}. Moreover, current contrastive frameworks lack a principle to prioritize hard-to-distinguish sample pairs, treating all pairs equally regardless of difficulty and limiting their effectiveness in highly imbalanced datasets.

To this end, we introduce \textbf{DyCON}, a \textbf{Dy}namic Uncertainty-aware \textbf{CON}sistency and Contrastive Learning framework designed to enhance consistency learning in semi-supervised medical image segmentation. DyCON introduces two specialized loss functions: Uncertainty-aware Consistency Loss (UnCL) and Focal Entropy-aware Contrastive Loss (FeCL). The former tackles global uncertainty by dynamically adjusting the contribution of each voxel to the consistency loss, while the latter refines local features to address subtle uncertainty and class imbalance.

Specifically, UnCL presents an uncertainty-aware consistency regularization that ensures the consistency loss captures information across all voxels rather than excluding high-uncertainty regions. By leveraging voxel-wise entropy from two networks (\eg teacher vs. student), UnCL dynamically scales the contribution of each voxel to the loss based on uncertainty. This adaptive mechanism enables the model to prioritize uncertain regions early in training, promoting exploration of challenging lesion boundaries without severe penalties. As training progresses, UnCL shifts its focus to refining predictions in confident regions, guiding DyCON to achieve stable and global consistency across the entire imaging context (see Fig.~\ref{fig:grad_cam} in Section~\ref{subsec:ablation}). 

While UnCL handles global adaptation to uncertainty, FeCL complements it by enhancing local feature discrimination with patch-wise contrastive learning. FeCL introduces dual focal mechanisms and adaptive confidence adjustment into the contrastive principle, where the similarity score of sample pairs are constrained based on their difficulty (hard positives and negatives) and uncertainty, respectively. This effectively addresses class imbalance by enhancing feature separability and enables DyCON to capture nuanced lesion boundaries.  % within challenging lesion patterns. 
Extensive experiments on four challenging datasets prove that our proposed DyCON significantly outperforms existing SOTA SSL methods (Fig.~\ref{fig:dycon_performance}). Our contributions are summarized as follows: 
\begin{itemize}
    % \item We introduce DyCON, a versatile semi-supervised framework tailored for medical image segmentation that enhances existing SSL methods without structural modifications, as shown in Fig.~\ref{fig:dycon_performance}.
    \item We introduce DyCON, a semi-supervised framework for medical image segmentation that dynamically integrates uncertainty awareness into both global and local learning to address class imbalance and lesion variations.
    % \item We propose UnCL, a global uncertainty-aware consistency loss, which scales voxel contributions based on entropy, thereby enabling DyCON to adaptively prioritize uncertain regions early in training and refine confident predictions as training progresses.
    \item We propose UnCL, a global uncertainty-aware consistency loss that scales voxel contributions by entropy to enable DyCON to prioritize uncertain regions initially and refine confident predictions over time.
    \item FeCL is proposed to enhance local discrimination in imbalanced and complex lesion settings by emphasizing hard positives/negatives and uncertain regions, guiding DyCON to capture subtle lesion boundaries.
    % \item Extensive experiments on four challenging medical image datasets (ISLES-2022, BraTS-2019, LA, Pancreas) validate DyCON's superior performance over SOTA SSL methods.
\end{itemize}
\vspace{-0.6em}
\section{Related Work}
\label{sec:related_work}
\textbf{Semi-supervised Medical Image Segmentation (SSMIS).} SSMIS aims to leverage unlabeled data to improve segmentation with limited annotations. The two widely used strategies are pseudo-labeling~\cite{bai:pseudo_labeling,wu:semi_pseudo_label,pseudo_label_threshold,chaitanya:local_contrastive,liu:prototype_pseudo_labeling,su:mutual_pseudo_labeling}, where unlabeled data are assigned pseudo-labels to augment training data, and consistency learning~\cite{ua_mt,mean_teacher,xu:ambiguity,xu:cross_modal_consistency,zhang:uncertainty_MCL,wang:mcf,li:diversity_mean_teacher,bai:BCP}, which encourages consistent predictions by leveraging diversity across different perturbations. Many SSMIS approaches combine these strategies. For instance, methods~\cite{self_loop_pseudo_label,yao:enhance_pseudo_label,wu:semi_pseudo_label,su:mutual_pseudo_labeling} incorporate pseudo-labels generated through self-training and cross-domain information. Recently, BCP~\cite{bai:BCP} and ABD~\cite{chi:ABD} aligns labeled and unlabeled distributions through bidirectional copy-pasting and displacement for robust pseudo-label generation, respectively. Consistency learning is achieved through various means, including co-training with multiple networks~\cite{wang:mcf,wu:cross_ml,xu:dual_uncertainty_guided,bai:BCP,chi:ABD}, mutual consistency with additional decoders~\cite{su:mutual_pseudo_labeling,gao:correlation_aware_co_training,wu:semi_pseudo_label}, and Mean-Teacher (MT) frameworks~\cite{mean_teacher,zhu:hybrid_double_uncertainty,zhang:multi_modal_mutual_learning}.
\begin{figure*}[!ht]
    \centering
    \includegraphics[width=16cm, height=7cm]{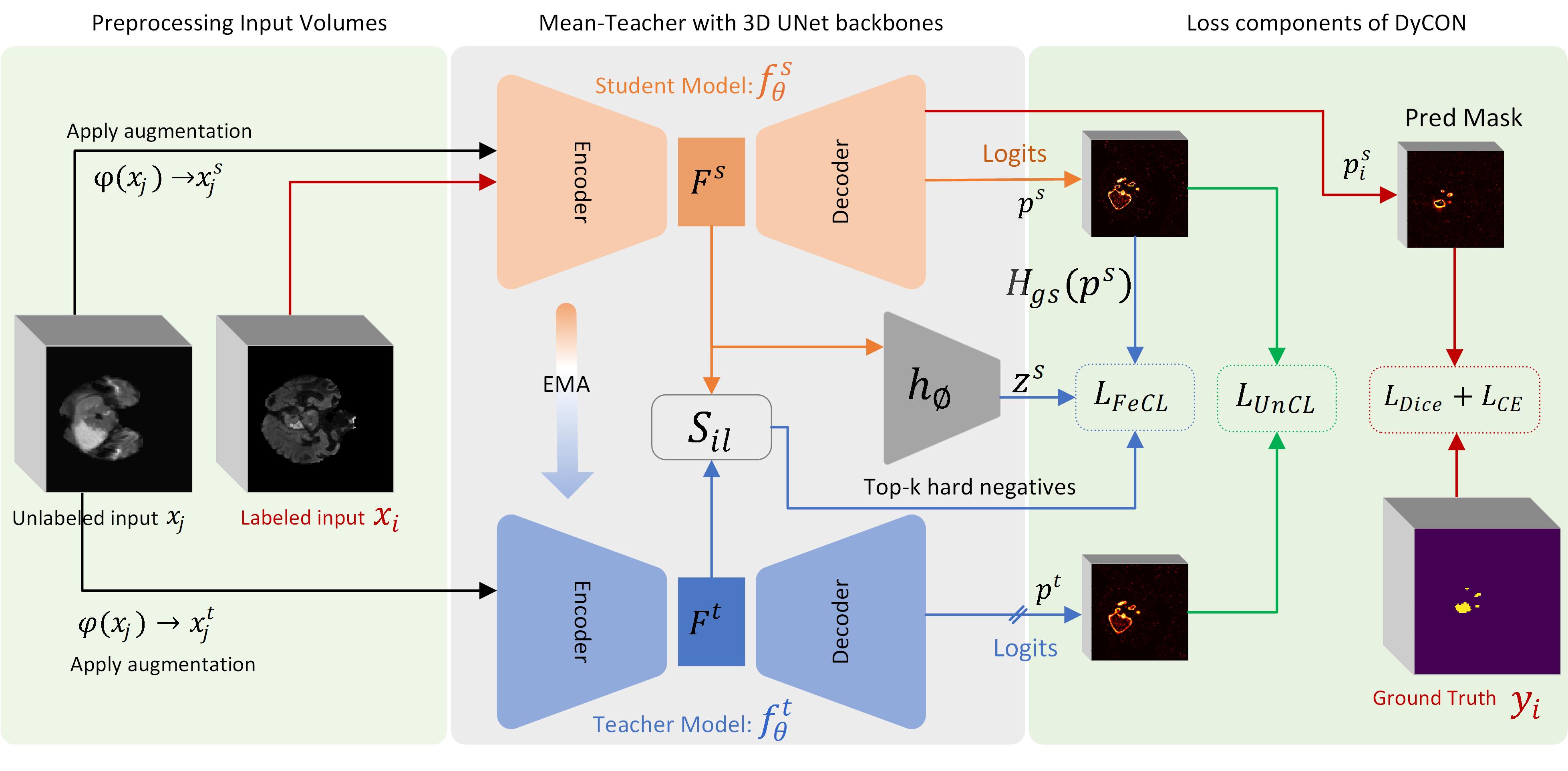}
    \caption{The overall framework of DyCON. Following the MT paradigm, the teacher model produces logits $p^t$ from unlabeled input $x_j^t$, while the student model generates logits $p^s$ and masks $p_i^s$ from unlabeled $x_j^s$ and labeled $x_i$ inputs, respectively. DyCON is ultimately optimized using the labeled ($\mathcal{L}_{Dice}$ and $\mathcal{L}_{CE}$) and unlabeled ($\mathcal{L}_{\text{UnCL}}$ and $\mathcal{L}_{\text{FeCL}}$) losses in an end-to-end semi-supervised manner. The FeCL loss uses patch embeddings ($z^s$) from a projection head ($h_\phi$) and lesion-wise hard negatives $\mathbf{S}_{il}$ to promote discriminative feature learning.}
    \label{fig:fig_DyCON}
    \vspace{-1.5em}
\end{figure*}

\textbf{Uncertainty Awareness in SSMIS.} In SSMIS, uncertainty arises from the inherent noise in pseudo-labels and the variance between model predictions. In response, uncertainty estimation enhances model robustness by informing consistency regularization and pseudo-labeling to focus on reliable regions~\cite{ct_lesion_uncertainty,xu:ambiguity,zhu:hybrid_double_uncertainty,zhang:uncertainty_MCL}. Current methods can be broadly grouped by their uncertainty estimation approaches. Model confidence methods utilize softmax probabilities variance to filter or weight predictions, thereby mitigating noise and refining labels based on confidence thresholds~\cite{yao:enhance_pseudo_label,su:mutual_pseudo_labeling,pseudo_label_threshold}. Bayesian approximations, such as Monte Carlo (MC) dropout, offer probabilistic uncertainty estimates through multiple stochastic forward passes, assisting target selection in noisy regions~\cite{zhu:hybrid_double_uncertainty,upc,fussnet_uncertainty,zhang:uncertainty_MCL,xu:ambiguity}. Deep ensemble approaches improve stability by aggregating multiple predictions to identify uncertain regions and selectively refining pseudo-labels~\cite{self_loop_pseudo_label,consistency_dyn_threshold}. Additionally, some recent methods leverage prediction discrepancies between dual-decoders~\cite{xu:dual_uncertainty_guided,zhang:multi_modal_mutual_learning} or sub-networks~\cite{wang:mcf,wu:cross_ml} as indicators of uncertainty, allowing models to focus on ambiguous regions for correction without excessive re-computation. In contrast, our proposed UnCL directly integrates uncertainty into a single-pass consistency loss as a self-contained measure, which dynamically down-weights high-uncertainty regions while preserving valuable information in moderately uncertain areas without requiring multiple predictions.

\textbf{Contrastive Learning in SSMIS.} Contrastive learning (CL)~\cite{supcon} learns discriminative features from unlabeled data by contrasting positive and negative sample pairs, guiding the model distinguish complex pathological structures. Recent progress in SSMIS has focused on adapting CL to the unique challenges of medical images. This includes incorporating local contrastive losses~\cite{chaitanya:local_contrastive,huang:dual_local_contrastive_consistency} to capture fine-grained details, using prototypes to represent diverse feature distributions~\cite{liu:prototype_pseudo_labeling}, and designing strategies to mine hard samples for enhanced discriminative power~\cite{you:ARCO,basak:PLGCL,tang:semi_hard_pos_contrast}. Furthermore, dual-stream learning and volume contrastive learning have been explored to improve feature discrimination and leverage contextual information~\cite{huang:dual_local_contrastive_consistency,Wu:VoCo}. 
In summary, while prior contrastive methods ensure feature consistency across augmentations or local contrasts, they treat all sample pairs equally, overlooking class imbalance and subtle lesion variations in brain segmentation. FeCL addresses this by combining focal weighting with entropy adjustments to emphasize hard-to-distinguish and uncertain regions for robust segmentation.
\vspace{-0.6em}
\section{Proposed Method}
\label{sec:proposed_method}
In this section, we present our UnCL and FeCL losses in DyCON framework. Although DyCON is flexible with any SSL frameworks, we integrate it with MT framework for simple presentation, as illustrated in Fig.~\ref{fig:fig_DyCON}. In SSMIS, let $\mathcal{X}=\{\mathcal{X}_l, \mathcal{X}_u\}$ denote a dataset of 3D MRI or CT images, split into labeled ($\mathcal{X}_l$) and unlabeled ($\mathcal{X}_u$) subsets. The labeled set $\mathcal{X}_l$ contains $\mathcal{N}$ samples, $\mathcal{X}_l = \{(x_i, y_i)\}_{i=1}^{\mathcal{N}}$, where $x_i \in \mathbb{R}^{H \times W \times D}$ is the 3D image, and $y_i \in \{0, 1, \dots, \mathcal{C}-1\}^{H \times W \times D}$ is the corresponding voxel-wise segmentation label (binary, with $\mathcal{C}=2$). The unlabeled set $\mathcal{X}_u$ has $\mathcal{K}$ samples, $\mathcal{X}_u = \{x\}_{j=\mathcal{N}+1}^{\mathcal{N}+\mathcal{K}}$, where $\mathcal{N}\ll\mathcal{K}$. Our objective is to leverage the limited labeled data $\mathcal{X}_l$ and abundant unlabeled data $\mathcal{X}_u$ to segment brain lesions in 3D images. To achieve this, we integrate DyCON with an MT architecture using two 3D UNet~\cite{3d_unet} models, $f_\theta^t$ and $f_\theta^s$, where $f_\theta^t$ is updated as the exponential moving average (EMA) of $f_\theta^s$. Additionally, an atrous spatial pyramid pooling (ASPP) module, followed by a convolutional layer, serves as a projection head $h_\phi(.)$ for contrastive learning. Finally, the DyCON loss $\mathcal{L}_{\text{DyCon}}$, combines supervised Dice ($\mathcal{L}_{\text{Dice}}$) and cross-entropy ($\mathcal{L}_{\text{CE}}$) losses (applied to labeled predictions $p_i^s$ against ground truth $y_i$) with unsupervised losses ($\mathcal{L}_{\text{UnCL}}$) and ($\mathcal{L}_{\text{FeCL}}$) for unlabeled data $\mathcal{X}_u$.
 
\subsection{UnCL: Uncertainty-aware Consistency Loss}
\label{subsec:beta_MSE}
UnCL introduces an uncertainty-aware consistency regularization that incorporates all voxels to guide the student model to learn from the teacher's confident predictions while allowing flexibility in uncertain areas. By leveraging entropy as an uncertainty, UnCL dynamically scales the voxel-wise consistency loss to achieve three goals: 1) Allowing exploration in ambiguous areas by reducing the penalty when both models show high uncertainty, 2) Amplifying the loss when one model is confident and the other uncertain, enforcing alignment with reliable predictions, and 3) Sharpening agreement in confident regions by strongly penalizing inconsistencies. As a result, this adaptive weighting enables effective knowledge transfer from teacher to student, enhancing segmentation accuracy.

Specifically, given two augmented views \(x_j^s\) and \(x_j^t\) of the same input \(x_j\), we obtain the prediction probabilities from the student model and teacher model as: $p_j^s = \sigma(f_\theta^s(x_j^s)),~p_j^t = \sigma(f_\theta^t(x_j^t))$, 
where \(\{p_j^s, p_j^t\} \in \mathbb{R}^{B \times H \times W \times D}\) and \(\sigma(.)\) denotes the softmax function. Note that the subscripts $j$ for predictions are \textbf{omitted henceforth for conciseness}. To align these predictions with uncertainty awareness, a consistency loss $\mathcal{L}(p^s, p^t)$, is weighted by a factor informed by the model's entropy across each voxel. The UnCL loss \(\mathcal{L}_{\text{UnCL}}\) is therefore defined as:
\begin{equation}
\vspace{-0.4em}
    \begin{aligned}
        \mathcal{L}_{\text{UnCL}} &= \frac{1}{N} \sum_{i=1}^{N} \frac{\mathcal{L}\left(p^s_i,p^t_i\right)}
        {\exp(\beta \cdot H_s(p^s_i)) + \exp(\beta \cdot H_t(p^t_i))} \\
        &\quad + \frac{\beta}{N} \sum_{i=1}^{N} \left(H_s(p^s_i) + H_t(p^t_i)\right)
    \end{aligned}
    \label{eq:l_UnCL}
    \vspace{-0.5em}
\end{equation}
where \(\mathcal{L}(p^s, p^t)\) is a distance-based loss function (e.g., MSE in this case) to align the student and teacher predictions. The entropy terms \(H_s(p^s)\) and \(H_t(p^t)\) respectively quantify the predictive uncertainty from student and teacher models. The hyperparameter \(\beta\) amplifies the degree of uncertainty. 

The consistency alignment is inversely weighted by the sum of \textbf{exponentiated entropies}, \(\exp(\beta \cdot H_s(p^s) + \beta \cdot H_t(p^t))\), derived from both student and teacher predictions. This exponential weighting modulates each voxel's influence based on prediction uncertainty, amplifying focus on confident regions (lower entropy) while reducing emphasis on uncertain areas (higher entropy). Intuitively, such weighting guides the model in prioritizing regions where both models are in agreement, while simultaneously scaling back the focus on regions where they diverge significantly. 

\begin{figure}[!t]
    \centering
    \includegraphics[width=8cm]{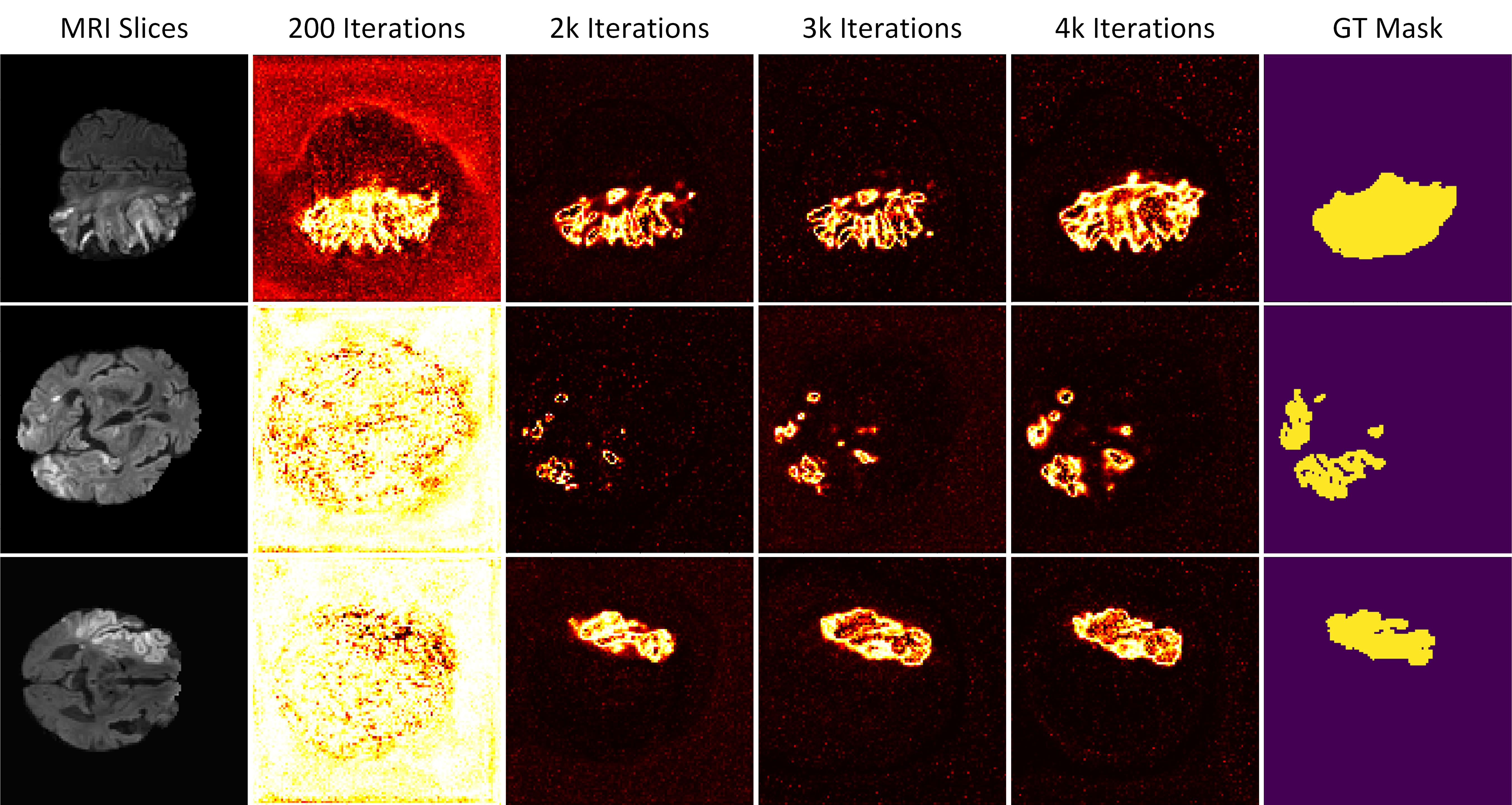}
    \vspace{-0.7em}
    \caption{Evolution of uncertainty maps with UnCL. Entropy maps over iterations show decreasing uncertainty, focusing around lesion regions and aligning closely with GT as training progresses.}
    \label{fig:entropy_maps}
    \vspace{-1em}
\end{figure}

To further encourage exploration without premature convergence, inspired by entropy minimization in UDA~\cite{vu:advent}, UnCL adds an \textbf{entropy regularization} term: \(\beta \cdot (H_s(p^s) + H_t(p^t))\) into the consistency training. This term acts as a counterbalance to the inversely weighted consistency loss. While the consistency loss focuses on aligning confident predictions, this term prevents the model from prematurely collapsing to overly confident (but potentially incorrect) predictions, especially in the early stages of training. Examples of estimated uncertainty for different lesions are illustrated in Fig.~\ref{fig:entropy_maps}.

The \textbf{adaptive $\beta$ parameter} serves as a scaling factor that amplifies or diminishes the impact of uncertainty on the loss for each voxel. During training, $\beta$ is dynamically adjusted, a higher $\beta$ value places greater emphasis on uncertain regions to encourage the model to focus its learning on those areas without being overly penalized early in training. Conversely, a lower $\beta$ value reduces the influence of uncertainty to allow the model to focus on refining its decisions in confident regions. The adaptive $\beta$ is defined as: $\beta(t) = \max(\beta_{\text{min}}, \beta_{\text{max}} \cdot \exp\left(-\lambda \cdot \frac{t}{T}\right))$, where, $\beta_{\text{max}}$ is the initial value of $\beta$, $\beta_{\text{min}}$ is the minimum value. $t$, $T$, and $\lambda$ denotes the current epoch, total epochs, and decay rate that controls how quickly $\beta$ decreases over time, respectively. This adaptive strategy ensures the model transitions smoothly from exploring uncertain areas to fine-tuning confident areas, accelerating convergence and improving segmentation quality (see gradient magnitudes in Fig.~\ref{fig:grad_magnitude}).

Moreover, the gradient analysis in Eq.(\ref{eq:grad_l_UnCL}) (more details in supplementary) reveals that UnCL effectively moderates the learning process by encouraging the student to learn from the teacher while actively handling uncertainties present in both models. This analysis confirms the desired properties of UnCL in promoting adaptive and robust learning for lesion segmentation. The gradient of $\mathcal{L}_{\text{UnCL}}$ w.r.t the student prediction $p^s$ is derived as:
\begin{equation}
% \vspace{-1em}
    \resizebox{\columnwidth}{!}{
        $\begin{aligned}
            \nabla_{p^s}\mathcal{L}_{\text{UnCL}} =
            &\ \frac{2(p^s - p^t)}{\exp(\beta \cdot H_s(p^s)) + \exp(\beta \cdot H_t(p^t))} \\
            & - \frac{(p^s - p^t)^2 \cdot \beta \cdot \frac{\partial H_s(p^s)}{\partial p^s} \cdot \exp(\beta \cdot H_s(p^s))}{\left(\exp(\beta \cdot H_s(p^s)) + \exp(\beta \cdot H_t(p^t))\right)^2}
            - \beta \cdot \frac{\partial H_s(p^s)}{\partial p^s}
        \end{aligned}$
    }
    \label{eq:grad_l_UnCL}
    % \vspace{-1em}
\end{equation}

\subsection{FeCL: Focal Entropy-aware Contrastive Loss}
\label{subsec:FoCL}
While UnCL enforces global consistency across voxel-wise predictions, yet it lacks the fine-grained handling of lesion details under class imbalance and intricate lesion distributions. Therefore, FeCL enforces patch-wise contrasts in the student model's feature space aiming to bring positive patch pairs (same class) closer and push negative pairs (different classes) apart. This guides the model to learn discriminative features that capture the subtle variations and contextual information.
To achieve this, FeCL introduces three mechanism: dual focal weighting, adaptive confidence adjustments, and inter-network top-$k$ hard negative mining.

In detail, intermediate features $F^s$ from the student's encoder are processed by a projection head $z^s=h_\phi(F^s)$ and partitioned into  $k\times k\times k$ patches. These patches are averaged, yielding compact representations $z_i^s \in \mathbb{R}^{\mathbb{N} \times h \times w \times d}$ that capture local context, where $i=\{1,2,\cdots,k^3\}$, $h=\frac{H}{k}$, $w=\frac{W}{k}$, $d=\frac{D}{k}$, and $\mathbb{N}=256$ is the embedding dimension. Similarly, the ground truth $y^s$ is partitioned and averaged, resulting in patched masks $y_i^s \in \mathbb{R}^{k^3}$, which are essential for defining positive and negative pairs in FeCL. Therefore, given patch embeddings $z_i^s$, the supervised contrastive loss~\cite{supcon} can be reformulated as FeCL:
\begin{equation}
    \resizebox{\columnwidth}{!}{
    $
    \begin{aligned}
        \mathcal{L}_{\text{FeCL}} &= \frac{1}{|P(i)|} \sum_{k \in P(i)} \mathbf{F}_k^+ \cdot 
        \left[ -\log \left( \frac{\exp(\mathbf{S}_{ik})}{D(i)} \right) \right] \\
         D(i) &= \exp(\mathbf{S}_{ik}) + \sum_{q \in N(i)} \mathbf{F}_q^- \cdot 
        \left[ \exp(\mathbf{S}_{iq}) + \frac{1}{K} \sum_{l=1}^{K} \exp(\mathbf{S}_{il}) \right]
    \end{aligned}
    $
    }
    \label{eq:combined}
\end{equation}
where $|P(i)|$ is the number of positive pairs for anchor $i$, $N(i)$ its negative pair set, $\mathbf{S}_{ik} = \frac{z_i^s \cdot z_k^s}{\tau}$ the similarity between $i$ and positive pair $k$, $\mathbf{S}_{iq}$ the similarity with negative pair $q$, and $\tau$ is a temperature parameter.

\textbf{Dual Focal Weights:} Instead of treating all positive and negative pairs equally, FeCL uses \textbf{patch-wise similarity scores} to identify hard positive and negative pairs: hard positives are same-class patches with low similarity, indicating areas where the model struggles to distinguish similarity, while hard negatives are different-class patches with high similarity, highlighting potential confusion. Therefore, to emphasize these challenging examples during the contrastive learning, FeCL applies dual focal weighting terms $\mathbf{F}_k^+$ and $\mathbf{F}_q^-$ in $\mathcal{L}_{\text{FeCL}}$ to adjust the importance of each patch pair in the loss which can be obtained as: 
\begin{equation}
    \mathbf{F}_k^+ = (1-\mathbf{S}_{ik})^\gamma \cdot \exp(H_{gs}(p_h^s)),~~\mathbf{F}_q^- = (\mathbf{S}_{iq})^\gamma
    \label{eq:f_+}
\end{equation}
The parameter $\gamma$ controls the focal weights and $H_{gs}$ represents the entropy from Gambling Softmax~\cite{jang:gambling_softmax} for student predictions $p^s$. Essentially, for positive pairs, the focal term $(1-\mathbf{S}_{ik})^\gamma$ enhances the importance of positive pairs with low similarity $\mathbf{S}_{ik}$, encouraging the model to align similar patches together. Conversely, for negative pairs, the focal term $(\mathbf{S}_{iq})^\gamma$ increases penalization for negative pairs with high similarity $\mathbf{S}_{iq}$, pushing apart dissimilar patches. Although FeCL shares mathematical elements with Focal Loss~\cite{lin:focal_loss}, its dual focal mechanism uniquely targets hard positive and negative pairs to refine feature representation in the embedding space, focusing on capturing subtle lesion details rather than classification accuracy.
 
FeCL further refines the focal weights $(1-\mathbf{S}_{ik})^\gamma$ for hard positives by incorporating entropy derived from Gambling Softmax, thereby it adaptively concentrates on uncertain or ambiguous boundaries. The adjusted probability $p_h^s$ for each voxel can be obtained using Gambling Softmax:
\begin{equation}
    p_h^s = \frac{\exp\left(\log(p^s) / T + (1 - \mathbb{C})\right)}{\sum_{k=1}^{\mathcal{C}} \exp\left(\log(p_k^s) / T + (1 - \mathbb{C})\right)}
    \label{eq:gs}
\end{equation} 
where $\mathbb{C}$ represents the confidence matrix: $\mathbb{C}=\max(p^s)$. Afterwards, the normalized entropy $H_{gs}(p_h^s)$ is used to weight the focal weights for positive pairs, so that patches with higher uncertainty have a larger impact on the loss, guiding the model towards these regions.

\textbf{Top-$k$ Hard Negatives:} The term \( \frac{1}{K} \sum_{l=1}^{K} \exp(\mathbf{S}_{il}) \) in FeCL captures the averaged impact of top-\(k\) hard negatives, enhancing the student model's ability to distinguish between similar yet incorrect examples. These hard negatives are selected from teacher embeddings \(F^t\) based on similarity to student embeddings \(F^s\), where \(\{i,l\}\in\{1,2,\dots,k^3\}\) are patches from both models.
\begin{equation}
    \begin{aligned}
    \mathbf{S}_{il} = \text{Top-}k\left( (F_i^s \cdot (F_l^t)^T) \odot \mathbf{M}_i \right)
    \label{eq:s_il}
    \end{aligned}
\end{equation}
where $K$ most similar negative patches that belong to a different class according to the ground truth \(\mathbf{M}_i\), are selected and incorporated into the denominator of the loss function.

\textbf{Total Loss Function:} The DyCON framework is trained end-to-end using a total loss $\mathcal{L}_{\text{Total}}$ that incorporates both supervised loss $\mathcal{L}_{\text{sup}}$ and unsupervised loss $\mathcal{L}_{\text{DyCon}}$, effectively using labeled and unlabeled data to improve performance. 
% \begin{equation}
%     \begin{aligned}
%         \mathcal{L}_{\text{DyCon}} = \mathcal{L}_{\text{Dice}} + \mathcal{L}_{\text{CE}} + \eta \cdot (\mathcal{L}_{\text{UnCL}} + \mathcal{L}_{\text{FeCL}}) 
%     \end{aligned}
%     \label{eq:l_total}
% \end{equation}
\begin{equation}
    \begin{aligned}
        \mathcal{L}_{\text{Total}} = 
        \underbrace{\mathcal{L}_{\text{Dice}} + \mathcal{L}_{\text{CE}}}_{\mathcal{L}_{\text{sup}}} + 
        \eta \cdot 
        \underbrace{(\mathcal{L}_{\text{UnCL}} + \mathcal{L}_{\text{FeCL}})}_{\mathcal{L}_{\text{DyCon}}}
    \end{aligned}
    \label{eq:l_total}
\end{equation}
where $\eta$ is a hyperparameter controlling the relative impact of the UnCL and FeCL losses in the total loss.
% \vspace{-0.7em}

\section{Experiments and Results}
\label{lab_experiments}
We extensively validate DyCON on diverse segmentation tasks, including brain lesions, tumors, and organs across MRI and CT, addressing challenges like varying shapes, sizes, and heterogeneous appearances.

\textbf{Dataset:} The \textbf{ISLES-2022} dataset~\cite{ISLES2022} contains 250 multi-channel MRI scans (DWI, ADC, FLAIR) for ischemic stroke lesion segmentation. We use only DWI for its sensitivity to acute lesions. Following supervised baselines~\cite{wu:w_net,feng:pamsnet}, we used 200 (80\%) images for training and 50 (20\%) images for testing and validation. The \textbf{BraTS2019} dataset~\cite{brats2019} contains 335 MRI scans with brain tumor annotations, where we focus on the T2-FLAIR sequence for whole tumor segmentation. With a split of 250/25/60 for training, validation, and testing. The \textbf{LA} dataset~\cite{LA_dataset} consists of 100 3D gadolinium-enhanced MRI scans of the left atrium, with a resolution of \(0.625\times 0.625\times 0.625mm^3\). 
The \textbf{Pancreas-NIH} dataset~\cite{Pancreas_dataset} contains 82 3D CT images with pancreas annotations, where its proximity to other organs and unclear boundaries make DyCON's uncertainty-handling ability crucial for precise boundary delineation. 

\textbf{Metrics: }We evaluate the segmentation performance of DyCON using Dice score, IoU, 95\% Hausdorff Distance (HD95), and Average Surface Distance (ASD), with all validations based on the student network's output.

\textbf{Implementation Details:}
All experiments are conducted on a single NVIDIA A100 80GB GPU using PyTorch framework. We follow standard practices from previous works~\cite{wang:mcf,xu:ambiguity,wu:cross_ml}, using 5\%, 10\% and 20\% labeled data, with the remainder as unlabeled, across all datasets. DyCON utilizes a 3D UNet~\cite{UNet} as the backbone, with an \textit{ASPP module} as the projection head \(h_\phi()\) for multi-scale feature extraction. Training images undergo 3D augmentations $\varphi(.)$, such as random cropping, flipping, and rotation, to improve generalization. Patches of size \(112 \times 112 \times 80\) are used for the LA dataset, \(96 \times 96 \times 96\) for Pancreas-NIH and BraTS2019, and \(96 \times 96 \times 64\) for ISLES-2022. The student network is optimized using \textit{SGD} with a learning rate of \(0.01\), momentum of 0.9, and weight decay of \(1 \times 10^{-4}\), trained for 6,000 iterations with batch size of 8 samples. Specifically, DyCON requires 8.58 hours on ISLES’22 with 5\% labels, while SOTA methods, such as BCP and GALoss respectively requires 7.52 and 9.2 hours. The loss weight \(\eta\) is set to 1.0, with \(\beta_{\text{max}}\)= 1.0, \(\beta_{\text{min}}\)= 0.1, and decay rate \(\lambda\)= 0.1 for ISLES-2022 and \(\lambda = 0.3\) for other datasets. The contrastive loss \(\tau\)=0.6, and patch sizes of \(k\)=16 is optimally used for all datasets to balance local and global context during training. The optimal focusing parameter $\gamma$=0.5 for ISLES'22 and $\gamma$=0.8 for other datasets.
% DyCON => ISLES-2022 Dataset Table
\begin{table}[!t]
    \centering
    \resizebox{\columnwidth}{!}{
    \begin{tabular}{lllllll}
        \toprule
        \multicolumn{1}{c}{\multirow{2}{*}{SSL Method}}  & \multicolumn{2}{c}{Volumes used in \textbf{ISLES'22}}       & \multicolumn{4}{c}{Metrics}        \\ \cmidrule{2-7} 
        \multicolumn{1}{c}{}        & Labeled                    & Unlabeled & Dice (\%)\(\uparrow\) & IoU (\%)\(\uparrow\) & HD95\(\downarrow\)  & ASD\(\downarrow\) \\ \midrule
        W-Net~\cite{wu:w_net}       & 200 (100\%)                & 0         & 85.60     & ---      & 27.34 & --- \\
        PAMSNet~\cite{feng:pamsnet} & 200 (100\%)                & 0         & 87.37     & 79.14    & 3.21  & --- \\ \midrule
        MT~\cite{mean_teacher}      &                            &           & 29.22     & 20.41    & 20.18 & 8.55  \\
        UA-MT~\cite{ua_mt}          &                            &           & 49.20     & 37.21    & 38.20 & 9.64 \\
        MCF~\cite{wang:mcf}         &                            &           & 39.79     & 29.83    & 40.67 & 10.65 \\
        CML~\cite{wu:cross_ml}      & \multirow{5}{*}{10 (5\%)}  & \multirow{5}{*}{190 (95\%)}      & 46.39     & 35.16    & 37.76 & 4.62 \\
        DTC~\cite{luo:DTC}          &                            &           & 46.55     & 34.80    & 37.33 & 8.18 \\
        AC-MT~\cite{xu:ambiguity}   &                            &           & 48.64     & 36.53    & 39.71 & 7.13 \\
        % DHC~\cite{dhc}              &                            &           &           &          &           \\
        MagicNet~\cite{magicnet}    &                            &           & 51.42     & 38.18    & 37.20 & 5.60  \\  
        GALoss~\cite{ga}            &                            &           & 53.29     & 40.17    & 31.72 & 4.53  \\ 
        BCP~\cite{bai:BCP}          &                            &           & 53.53     & 41.12    & 37.06 & 6.91 \\
        \textbf{DyCON (Ours)}       &                            &           & \textbf{61.48}       & \textbf{48.80}  & \textbf{17.61}  & \textbf{0.75} 
        \\ \midrule
        MT~\cite{mean_teacher}      &                            &           & 36.43     & 24.01    & 21.80 & 7.22  \\
        MCF~\cite{wang:mcf}         &                            &           & 42.96     & 32.51    & 42.82 & 10.86 \\
        DTC~\cite{luo:DTC} & \multirow{5}{*}{20 (10\%)} & \multirow{5}{*}{180 (90\%)}    & 45.19    & 32.80 & 36.24  & 5.10 \\
        AC-MT~\cite{xu:ambiguity}   &                            &           & 49.47     & 37.02    & 39.67 & 11.10 \\
        CML~\cite{wu:cross_ml}      &                            &           & 50.88     & 38.45    & 36.16 & 4.94 \\
        BCP~\cite{bai:BCP}          &                            &           & 57.97     & 44.32    & 30.09 & 4.58 \\
        % DHC~\cite{dhc}              &                            &           &           &          &           \\ 
        MagicNet~\cite{magicnet}    &                            &           & 58.84     & 44.42    & 29.18 & 3.64  \\  
        GALoss~\cite{ga}            &                            &           & 60.13     & 47.27    & 24.11 & 3.17  \\  
        \textbf{DyCON (Ours)}       &                            &           & \textbf{65.71}       & \textbf{51.09} & \textbf{13.35}   & \textbf{0.71} \\ 
        \midrule
        MT~\cite{mean_teacher}      &                            &           & 37.70     & 26.33    & 19.00 & 6.45  \\
        UA-MT~\cite{ua_mt}          &                            &           & 58.00     & 44.96    & 28.99 & 3.13  \\
        DTC~\cite{luo:DTC} & \multirow{3}{*}{40 (20\%)} & \multirow{3}{*}{160 (80\%)}    & 40.23    & 29.35 & 41.47  & 13.13 \\
        MCF~\cite{wang:mcf}         &                            &           & 40.36     & 31.31    & 41.10 & 13.03 \\
        AC-MT~\cite{xu:ambiguity}   &                            &           & 54.91     & 41.55    & 32.27 & 2.36 \\
        CML~\cite{wu:cross_ml}      &                            &           & 54.31     & 41.77    & 30.75 & 1.35 \\
        BCP~\cite{bai:BCP}          &                            &           & 60.35     & 46.41    & 29.63 & 3.64 \\
        % \midrule
        % % DHC~\cite{dhc}              &                            &           &           &          &           \\
        % MagicNet~\cite{magicnet}    &                            &           &           &          &           \\        
        % GALoss~\cite{ga}            &                            &           &           &          &           \\
        \textbf{DyCON (Ours)}       &                            &           & \textbf{69.11}      & \textbf{54.74} & \textbf{10.58}   & \textbf{0.52} \\
        \bottomrule
    \end{tabular}
    }    
    \vspace{-0.6em}
    \caption{Performance comparison with SOTA methods on ISLES'22 with various label proportions. The results indicate that DyCON attains superior performance regardless of label settings. }
    \label{tab:isles_table}
    \vspace{-1.5em}
\end{table}
\vspace{-1.5em}
\subsection{Comparison with State-of-the-art Methods}
\label{subseq:comparison_results}
This section benchmarks DyCON against SOTA SSL methods, demonstrating its effectiveness in handling uncertainty with limited supervision. All best results are averaged over 3-fold cross-validation for a fair and consistent evaluation.

\textbf{ISLES-2022:} To our knowledge, DyCON is the first to apply SSL for stroke lesion segmentation on the ISLES'22 dataset, which presents significant variability in lesion size, location, and sparsely distributed lesions. For a fair comparison, we meticulously reproduced the results of SOTA methods on ISLES'22 using the same backbone and experimental setup as DyCON. As shown in Table \ref{tab:isles_table}, despite the inherent challenges in this dataset, DyCON consistently outperforms all methods across various metrics and label proportions. For instance, with 5\% labeled data, DyCON reaches a Dice score of 61.48\%, outperforming BCP's 53.53\%. Similarly, DyCON notably surpasses the recent class-imbalanced methods: GALoss~\cite{ga} (\(65.71\%\) vs. \(60.13\%\) Dice) using 10\% labels of this challenging dataset.

\begin{figure}[!ht]
    \centering
    \vspace{-0.5em}
    \includegraphics[width=8.2cm, height=5cm]{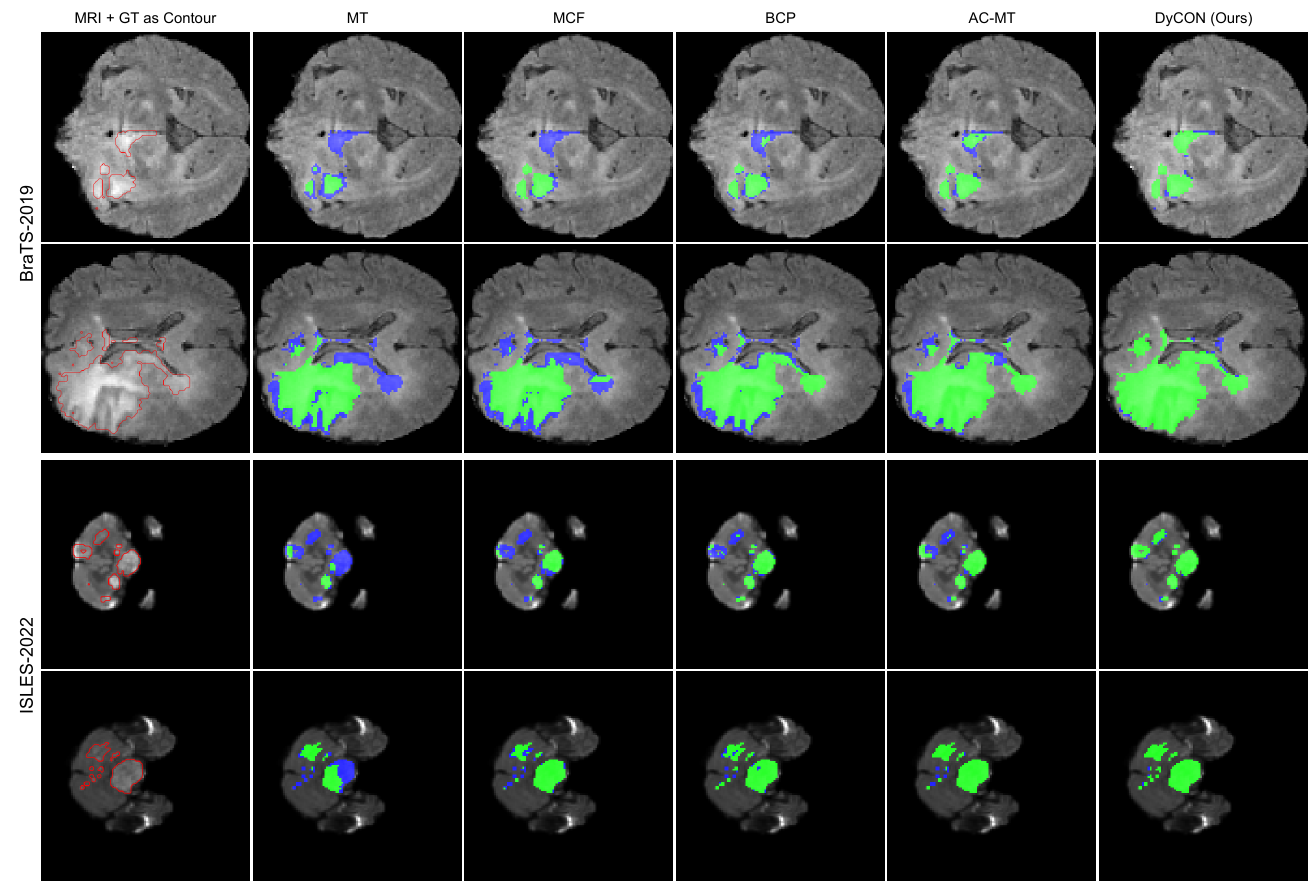}
    \vspace{-0.5em}
    \caption{Lesion and Tumor segmentation visualization on ISLES'22 and BraTS'19 datasets with 10\% labeled data.
    }
    \label{fig:seg_brats19_isles22}
    \vspace{-0.9em}
\end{figure}

\textbf{BraTS-2019:} DyCON also exhibits strong performance on the BraTS2019 dataset, known for its class imbalance and heterogeneous nature of tumors (Table~\ref{tab:brats19_table}). With 10\% and 20\% labeled data, DyCON achieves a Dice score of 87.05\% and 88.75\%, outperforming the recent CML, which produced 85.26\% and 86.63\%, respectively. Fig.~\ref{fig:seg_brats19_isles22} further indicates DyCON's superior performance in segmenting lesions and tumors on ISLES'22 and BraTS'19 with only 10\% labeled data. DyCON captures small and scattered lesions (which are often the hardest to detect) with more true positives and fewer false negatives than other SSL methods.
% DyCON => BraTS-2019 Dataset Table
\begin{table}[!t]
    \centering
    \resizebox{\columnwidth}{!}{
    \begin{tabular}{lllllll}
        \toprule
        \multicolumn{1}{c}{\multirow{2}{*}{SSL Method}}  & \multicolumn{2}{c}{Volumes used in \textbf{BraTS'19}}       & \multicolumn{4}{c}{Metrics}        \\ \cmidrule{2-7} 
        \multicolumn{1}{c}{}        & Labeled                    & Unlabeled & Dice (\%)\(\uparrow\) & IoU (\%)\(\uparrow\) & HD95\(\downarrow\)  & ASD\(\downarrow\) \\ \midrule
        3D-UNet~\cite{UNet}           & 250 (100\%)                 & 0         & 88.23     & 78.81   & 7.21  & 1.53 \\ 
        \midrule
        MT~\cite{mean_teacher}      &                            &           & 81.70     & 70.82    & 22.29 & 7.36   \\
        URPC~\cite{LU:urpc}         &                            &           & 74.59     & 63.11    & 13.88 & 3.72   \\
        UA-MT~\cite{ua_mt}          &                            &           & 82.82     & 72.77    & 11.29 & 2.30 \\
        DTC~\cite{luo:DTC}          & \multirow{4}{*}{25 (10\%)} & \multirow{4}{*}{225 (90\%)} & 81.57& 71.63 & 15.73  & 2.56 \\
        MCF~\cite{wang:mcf}         &                            &           & 83.67     & 72.15    & 12.58 & 3.28    \\
        BCP~\cite{bai:BCP}          &                            &           & 83.42     & 73.31    & 10.11 & 1.89    \\
        AC-MT~\cite{xu:ambiguity}   &                            &           & 83.77     & 73.96    & 11.35 & 1.93    \\
        CML~\cite{wu:cross_ml}      &                            &           & 85.26     & ---      & 9.08  & 1.83\\
        % ABD~\cite{chi:ABD}          &                            &           &           &          &       &     \\
        \textbf{DyCON (Ours)}       &                            &           & \textbf{87.05}       & \textbf{77.73}   & \textbf{7.41}  & \textbf{1.14} 
        \\ \midrule
        MT~\cite{mean_teacher}      &                            &           & 83.04     & 72.10    & 9.85 & 2.32   \\
        URPC~\cite{LU:urpc}         &                            &           & 82.93     & 72.57    & 5.93 & 3.19   \\
        UA-MT~\cite{ua_mt}          &                            &           & 83.61     & 73.98    & 11.44 & 2.26 \\
        DTC~\cite{luo:DTC}          & \multirow{4}{*}{50 (20\%)} & \multirow{4}{*}{200 (80\%)} & 83.43 & 73.56 & 14.77  & 2.34 \\
        MCF~\cite{wang:mcf}         &                            &           & 84.85     & 73.61    & 11.24 & 2.29 \\
        BCP~\cite{bai:BCP}          &                            &           & 82.71     & 72.72    & 9.99 & 1.86 \\
        AC-MT~\cite{xu:ambiguity}   &                            &           & 84.63     & 74.39    & 9.50 & 2.11 \\
        CML~\cite{wu:cross_ml}      &                            &           & 86.63     & ---      & 7.83 & 1.45\\
        % ABD~\cite{chi:ABD}          &                            &           &           &          &       &     \\
        \textbf{DyCON (Ours)}       &                            &           & \textbf{88.75}       & \textbf{80.52}         & \textbf{6.33}      & \textbf{0.93} \\
        \bottomrule
    \end{tabular}
    }    \vspace{-0.6em}
    \caption{Comparison of tumor segmentation performance with SOTA methods on BraTS'19 Dataset. Notably, DyCON outperforms all methods with a significant margin across all metrics.}
    \label{tab:brats19_table}
    \vspace{-1.5em}
\end{table}

\vspace{-1.2em}
\textbf{LA and Pancreas:} DyCON further demonstrates consistent performance improvement across diverse datasets that underscores its generalizability. On the LA dataset (Table \ref{tab:la_table}), DyCON respectively with 3D-UNet and VNet attains a Dice score of 90.96\% and 91.18\% with just 5\% labeled data, showing its backbone-agnostic nature. Similarly, on PancreasCT (Table \ref{tab:pancreas_table}), it reaches a Dice score of 84.81\% with 20\% labeled data. In conclusion, DyCON’s improved performance across diverse tasks (modalities) highlights its strength in generalization by handling voxel-wise uncertainty and class imbalance. 

\subsection{Ablation Study}
\label{subsec:ablation}
 We perform ablation studies to demonstrate the impact of each component in DyCON using ISLES'22 and BraTS2019 datasets with 10\% and 20\% label settings. While DyCON is flexible with any SSL framework, we integrate it with MT framework as a baseline for consistent evaluation across all experiments. 
% DyCON => LA Dataset Table
\begin{table}[!t]
    \centering
    \resizebox{\columnwidth}{!}{
    \begin{tabular}{lllllll}
        \toprule
        \multicolumn{1}{c}{\multirow{2}{*}{SSL Method}}  & \multicolumn{2}{c}{Volumes used in \textbf{LA}}       & \multicolumn{4}{c}{Metrics}        \\ \cmidrule{2-7} 
        \multicolumn{1}{c}{}        & Labeled                    & Unlabeled & Dice (\%)\(\uparrow\) & IoU (\%)\(\uparrow\) & HD95\(\downarrow\)  & ASD\(\downarrow\) \\ \midrule
        % W-Net~\cite{wu:w_net}       & 80 (100\%)                 & 0         & 85.60     & ---      & 27.34 & --- \\
        3D-UNet~\cite{3d_unet}                       & 80 (100\%)                 & 0         & 91.51     & 84.04    & 1.53  & 5.61 \\ 
        \midrule
        UPC~\cite{upc}              &                            &           & 86.36     & 76.24    & 13.83 & 3.64   \\
        UA-MT~\cite{ua_mt}          &                            &           & 88.34     & 76.11    & 10.01 & 4.43   \\
        DTC~\cite{luo:DTC}          & \multirow{5}{*}{4 (5\%)}  & \multirow{5}{*}{76 (95\%)} & 81.25& 74.26 & 14.90  & 3.99    \\
        MCF~\cite{wang:mcf}         &                            &           & 86.52     & 77.43    & 9.12  & 2.40    \\
        BCP~\cite{bai:BCP}          &                            &           & 88.02     & 78.72    & 7.90  & 2.15    \\
        CML~\cite{wu:cross_ml}      &                            &           & 87.63     & ---      & 8.92  & 2.23    \\
        AC-MT~\cite{xu:ambiguity}   &                            &           & 89.12     & 80.46    & 11.05 & 2.19    \\
        DyCON (3D-UNet)       &                            &           & 90.96       & 83.54   & 5.39  & 1.91 \\ 
        \textbf{DyCON (VNet)}       &                            &           & \textbf{91.18}       & \textbf{84.16}   & \textbf{5.16}  & \textbf{1.39} \\ 
        \midrule
        UPC~\cite{upc}              &                            &           & 89.65     & 81.36    & 6.71 & 2.15   \\
        UA-MT~\cite{ua_mt}          &                            &           & 90.16     & 82.18    & 6.50 & 1.98 \\
        DTC~\cite{luo:DTC}          & \multirow{5}{*}{8 (10\%)} & \multirow{5}{*}{72 (90\%)} & 87.51& 78.17& 8.23  & 2.36 \\
        MCF~\cite{wang:mcf}         &                            &           & 88.71     & 80.41    & 6.32 & 1.90 \\
        BCP~\cite{bai:BCP}          &                            &           & 89.62     & 81.31    & 6.81 & 1.76 \\
        AC-MT~\cite{xu:ambiguity}   &                            &           & 90.31     & 82.43    & 6.21 & 1.76\\
        CML~\cite{wu:cross_ml}      &                            &           & 90.36     & ---      & 6.06 & 1.68 \\
        DyCON (VNet)       &                            &           & 91.58       & 84.40   & 5.02  & 1.52 \\ 
        \textbf{DyCON (3D-UNet)}       &                            &           & \textbf{92.77}       & \textbf{86.21} & \textbf{4.20}  & \textbf{1.23} \\
        \bottomrule
    \end{tabular}
    }    \vspace{-0.5em}
    \caption{Comparison of segmentation performance on Left Atrium dataset with SOTA methods using 5\% and 10\% label regimes.  }
    \label{tab:la_table}
    \vspace{-0.6em}
\end{table}

% DyCON => Pancreas CT Dataset Table
\begin{table}[!t]
    \centering
    \resizebox{\columnwidth}{!}{
    \begin{tabular}{lllllll}
        \toprule
        \multicolumn{1}{c}{\multirow{2}{*}{SSL Method}}  & \multicolumn{2}{c}{Volumes used in \textbf{Pancreas CT}}       & \multicolumn{4}{c}{Metrics}        \\ \cmidrule{2-7} 
        \multicolumn{1}{c}{}        & Labeled                    & Unlabeled & Dice (\%)\(\uparrow\) & IoU (\%)\(\uparrow\) & HD95\(\downarrow\)  & ASD\(\downarrow\) \\ \midrule
        V-Net~\cite{UNet}           & 62 (100\%)                 & 0         & 69.95     & 55.56    & 14.23  & 1.64 \\ 
        \midrule
        MT~\cite{mean_teacher}      &                            &           & 71.43     & 60.21    & 15.44 & 4.11   \\
        UA-MT~\cite{ua_mt}          &                            &           & 77.26     & 63.82    & 11.90 & 3.06   \\
        DTC~\cite{luo:DTC}          & \multirow{2}{*}{12 (20\%)} & \multirow{2}{*}{50 (80\%)} & 78.27& 64.75& 8.36  & 2.25    \\
        MCF~\cite{wang:mcf}         &                            &           & 75.00     & 61.27    & 11.59 & 3.27    \\
        CML~\cite{wu:cross_ml}      &                            &           & 77.26     & 56.21    & 25.82 & 1.52    \\
        BCP~\cite{bai:BCP}          &                            &           & 82.91     & 70.97    & 6.43  & 2.25    \\
        % ABD~\cite{chi:ABD}          &                            &           &           &          &       &         \\
        % PLGCL~\cite{basak:PLGCL}    &                            &           &           &          &       &         \\
        \textbf{DyCON (Ours)}       &                            &           & \textbf{84.81}  & \textbf{73.86}  & \textbf{5.41}  & \textbf{1.44} \\
        \bottomrule
    \end{tabular}
    }    \vspace{-0.8em}
    \caption{Comparison of segmentation performance on Pancreas CT dataset with SOTA methods using 20\% labeled data.}
    \label{tab:pancreas_table}
    \vspace{-1.5em}
\end{table}
 
\textbf{Effectiveness of UnCL for Uncertainty Estimation:}
We investigate the role of entropy as a dynamic measure of uncertainty in the MT training process. This experiment evaluates two configurations: (1) Vanilla MT, (2) MT+UnCL without a scaling factor ($\beta$=\(\emptyset\)) and (3) MT+UnCL using fixed and adaptive $\beta$ values. Note that FeCL is not considered in this experiment. The results in Table~\ref{tab:abl_study} clearly demonstrate the effectiveness of UnCL in improving segmentation performance. We observe that incorporating entropy weighting, even without $\beta$ yields substantial improvements over the baseline MT. Furthermore, increasing $\beta$ values led to consistent performance gains, demonstrating the benefit of amplifying the focus on uncertain regions. Notably, the adaptive $\beta$ strategy leads to best performance across all scenarios in both datasets. 
\begin{table}[!t]
    \centering
    \resizebox{\columnwidth}{!}{
        \begin{tabular}{ccclllllll}
            \toprule
            \multicolumn{2}{c}{Scans used}  & \multicolumn{1}{c}{\multirow{2}{*}{UnCL}} & \multicolumn{1}{c}{\multirow{2}{*}{$\beta$}} & \multicolumn{3}{c}{ISLES-2022} & \multicolumn{3}{c}{BraTS-2019} \\ \cmidrule{1-2} \cmidrule{5-10} 
            Labeled    & Unlabeled  & \multicolumn{1}{c}{}   & \multicolumn{1}{c}{}  & Dice(\%)\(\uparrow\)  & HD95\(\downarrow\)  & ASD\(\downarrow\)  & Dice(\%)\(\uparrow\) & HD95\(\downarrow\)  & ASD\(\downarrow\)  \\ 
            \midrule
            \multirow{5}{*}{10\%} & \multirow{5}{*}{90\%} & \ding{55}  & \ding{55} & 36.43 & 21.80 & 7.22 & 81.70 & 22.29 & 7.36   \\
                                  % &                       & \ding{55}  & \ding{55} &   &   &   &   &   &         \\
                                  &                       & \checkmark & \ding{55} & 58.78 & 19.37 & 6.16 & 80.04  & 14.35 & 4.23    \\
                                  &                       & \checkmark & 0.5       & 60.97 & 16.89 & 5.01 & 83.11  & 12.23 & 3.00    \\
                                  &                       & \checkmark & 0.8       & 62.23 & 15.18 & 3.42 & 84.12  & 10.1  & 2.31    \\
                                  &                       & \checkmark & $\top$    & \textbf{64.52} & \textbf{14.10} & \textbf{1.05} & \textbf{85.97}  & \textbf{8.50}  & \textbf{1.78}    \\ 
                                  \midrule
            \multirow{5}{*}{20\%} & \multirow{5}{*}{80\%} & \ding{55}  & \ding{55} & 37.70  & 19.00 & 6.45 & 83.04 & 9.85  & 2.32    \\
                                  % &                       & \ding{55}  & \ding{55} &    &   &   &   &   &         \\
                                  &                       & \checkmark & \ding{55} & 60.68  & 15.76 & 3.11 & 81.23  & 12.14 & 3.20   \\
                                  &                       & \checkmark & 0.5       & 63.15  & 14.51 & 2.13 & 83.68  & 10.03 & 2.74   \\
                                  &                       & \checkmark & 0.8       & 66.05  & 12.64 & 1.20 & 85.77  & 8.99  & 1.86   \\
                                  &                       & \checkmark & $\top$    & \textbf{68.30}  & \textbf{11.12} &\textbf{ 0.96} & \textbf{87.03}  & \textbf{7.18}  & \textbf{1.24}   \\
            \bottomrule
        \end{tabular}
    }    \vspace{-0.5em}
    \caption{Ablation results on different UnCL configurations, including UnCL without $\beta$ and with fixed/adaptive $\beta$ values on the ISLES'22 and BraTS'19 datasets using 10\% and 20\% labeled data.}
    \label{tab:abl_study}
    \vspace{-1em}
\end{table}

\begin{figure}[!t]
    \centering
    % \vspace{-0.5em}
    \includegraphics[width=8cm]{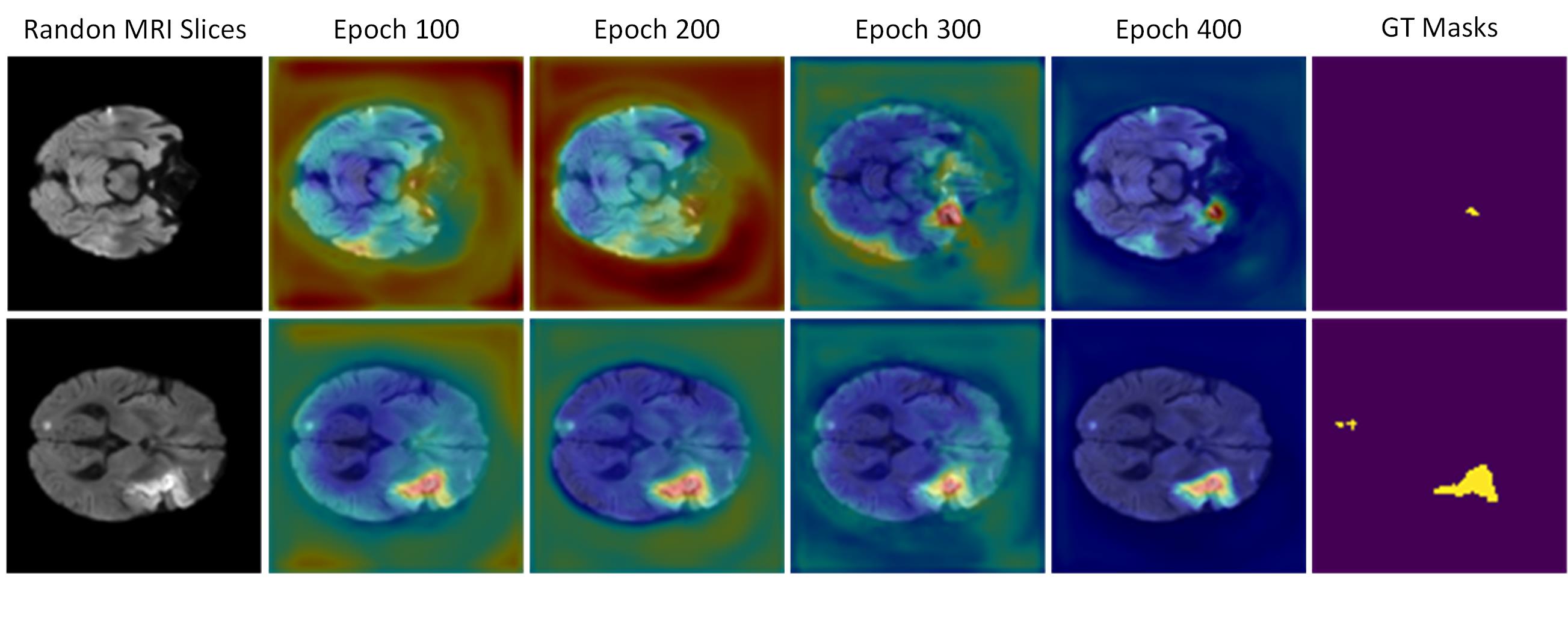}
    \vspace{-1em}
    \caption{Grad-CAM over training epochs shows the model progressively focuses on lesion regions aligning with ground truth.}
    \label{fig:grad_cam}
    \vspace{-1.4em}
\end{figure}

Fig.~\ref{fig:grad_magnitude} illustrates how $\beta$ values affect gradient magnitudes. Higher $\beta$ values (red, teal) increase initial gradients, focusing on uncertain regions, while lower values (blue, orange) provide gradual focus. The adaptive $\beta$ (red) balances this over time consistently across ISLES'22 and BraTS'19. The Grad-CAM visualization in Fig.~\ref{fig:grad_cam} further highlights the focus of DyCON sharpening over training from broad activations (epochs 100) to precise lesion localization (epochs 400), aligning with gradient analysis and underscoring the role of UnCL in enhancing segmentation accuracy.
\definecolor{orange_ryb}{rgb}{0.98, 0.6, 0.01}
\definecolor{bleudefrance}{rgb}{0.19, 0.55, 0.91}
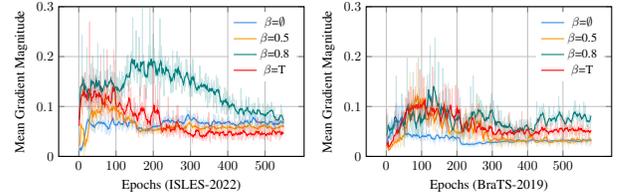
\begin{figure}[!t]
    \centering
    % First subfigure (left)
    \begin{subfigure}{0.48\linewidth}  % Adjust width for two-column layout
        \centering
        % \caption{Plot 1}
        \resizebox{\columnwidth}{!}{
            \begin{tikzpicture}
                \begin{axis}[
                    width=8cm,
                    height=5.5cm,
                    xlabel={Epochs (ISLES-2022)},
                    ylabel={Mean Gradient Magnitude},
                    legend pos=north east,
                    legend style={draw=none, font=\small},
                    grid=major,
                    ymin=0, ymax=0.3,  % Adjust the y-axis range to match your plot
                    xtick={0,100,200,300,400,500}, % Adjust as needed for the x-ticks
                    thick,
                ]
                % Read data from CSV file
                \pgfplotstableread[col sep=comma]{figures/grad_original_and_smoothed_isles22.csv}\datatable
                % Plot smoothed gradients
                \addplot[bleudefrance, thick] table [x=Epoch, y=beta_nll_s] {\datatable};
                \addlegendentry{$\beta$=$\emptyset$}
                \addplot[orange_ryb, thick] table [x=Epoch, y=beta_05_s] {\datatable};
                \addlegendentry{$\beta$=0.5}
                \addplot[teal, thick] table [x=Epoch, y=beta_08_s] {\datatable};
                \addlegendentry{$\beta$=0.8}
                \addplot[red, thick] table [x=Epoch, y=beta_T_s] {\datatable};
                \addlegendentry{$\beta$=T}
                % Plot original gradients with transparency (opacity)
                \addplot[bleudefrance, thin, opacity=0.2] table [x=Epoch, y=beta_nll] {\datatable};
                \addplot[orange_ryb, thin, opacity=0.2] table [x=Epoch, y=beta_05] {\datatable};
                \addplot[teal, thin, opacity=0.2] table [x=Epoch, y=beta_08] {\datatable};
                \addplot[red, thin, opacity=0.2] table [x=Epoch, y=beta_T] {\datatable};
                \end{axis}
            \end{tikzpicture}
        }
    \end{subfigure}
    % Second subfigure (right)
    \begin{subfigure}{0.48\linewidth}
        \centering
        % \caption{Plot 2}
        \resizebox{\columnwidth}{!}{
            \begin{tikzpicture}
                \begin{axis}[
                    width=8cm,
                    height=5.5cm,
                    xlabel={Epochs (BraTS-2019)},
                    ylabel={Mean Gradient Magnitude},
                    legend pos=north east,
                    legend style={draw=none, font=\small},
                    grid=major,
                    ymin=0, ymax=0.3,  % Adjust the y-axis range to match your plot
                    xtick={0,100,200,300,400,500}, % Adjust as needed for the x-ticks
                    thick,
                ]
                % Read data from CSV file
                \pgfplotstableread[col sep=comma]{figures/grad_original_and_smoothed_brats19.csv}\datatable
                % Plot smoothed gradients
                \addplot[bleudefrance, thick] table [x=Epoch, y=beta_nll_s] {\datatable};
                \addlegendentry{$\beta$=$\emptyset$}
                \addplot[orange_ryb, thick] table [x=Epoch, y=beta_05_s] {\datatable};
                \addlegendentry{$\beta$=0.5}
                \addplot[teal, thick] table [x=Epoch, y=beta_08_s] {\datatable};
                \addlegendentry{$\beta$=0.8}
                \addplot[red, thick] table [x=Epoch, y=beta_T_s] {\datatable};
                \addlegendentry{$\beta$=T}
                % Plot original gradients with transparency (opacity)
                \addplot[bleudefrance, thin, opacity=0.2] table [x=Epoch, y=beta_nll] {\datatable};
                \addplot[orange_ryb, thin, opacity=0.2] table [x=Epoch, y=beta_05] {\datatable};
                \addplot[teal, thin, opacity=0.2] table [x=Epoch, y=beta_08] {\datatable};
                \addplot[red, thin, opacity=0.2] table [x=Epoch, y=beta_T] {\datatable};
                \end{axis}
            \end{tikzpicture}
        }
    \end{subfigure}
    % \vspace{-1em}
        \caption{Batch-wise mean gradient magnitudes for different $\beta$ values over time. Solid lines show smoothed trends, while transparent lines capture original gradient variability. $T$ refer to adaptively changing values of $\beta$.}
    \label{fig:grad_magnitude}
    \vspace{-1em}
\end{figure}

\textbf{Effectiveness of FeCL for Class Imbalance:}
To evaluate each component of FeCL loss, we isolate the effects of focal mechanisms (F$^+$+F$^-$), top-$k$ hard negatives (HN), and entropy-awareness (Entropy). Since FeCL builds on SupCon~\cite{supcon}, we integrate both losses into MT frameworks and test MT+SupCon and MT+FeCL on both datasets using 10\% labeled data. This analysis highlights role of FeCL in improving model's capability to distinguish challenging regions where traditional methods struggle. Finally, the joint effects of UnCL and FeCL in DyCON (MT+UnCL+FeCL) is also evaluated, as reported in Table~\ref{tab:FeCL_results}. 

\begin{figure}[!t]
    \centering
    % \vspace{-1em}
    \begin{subfigure}[t]{0.46\linewidth}
        \centering
        \includegraphics[width=\linewidth]{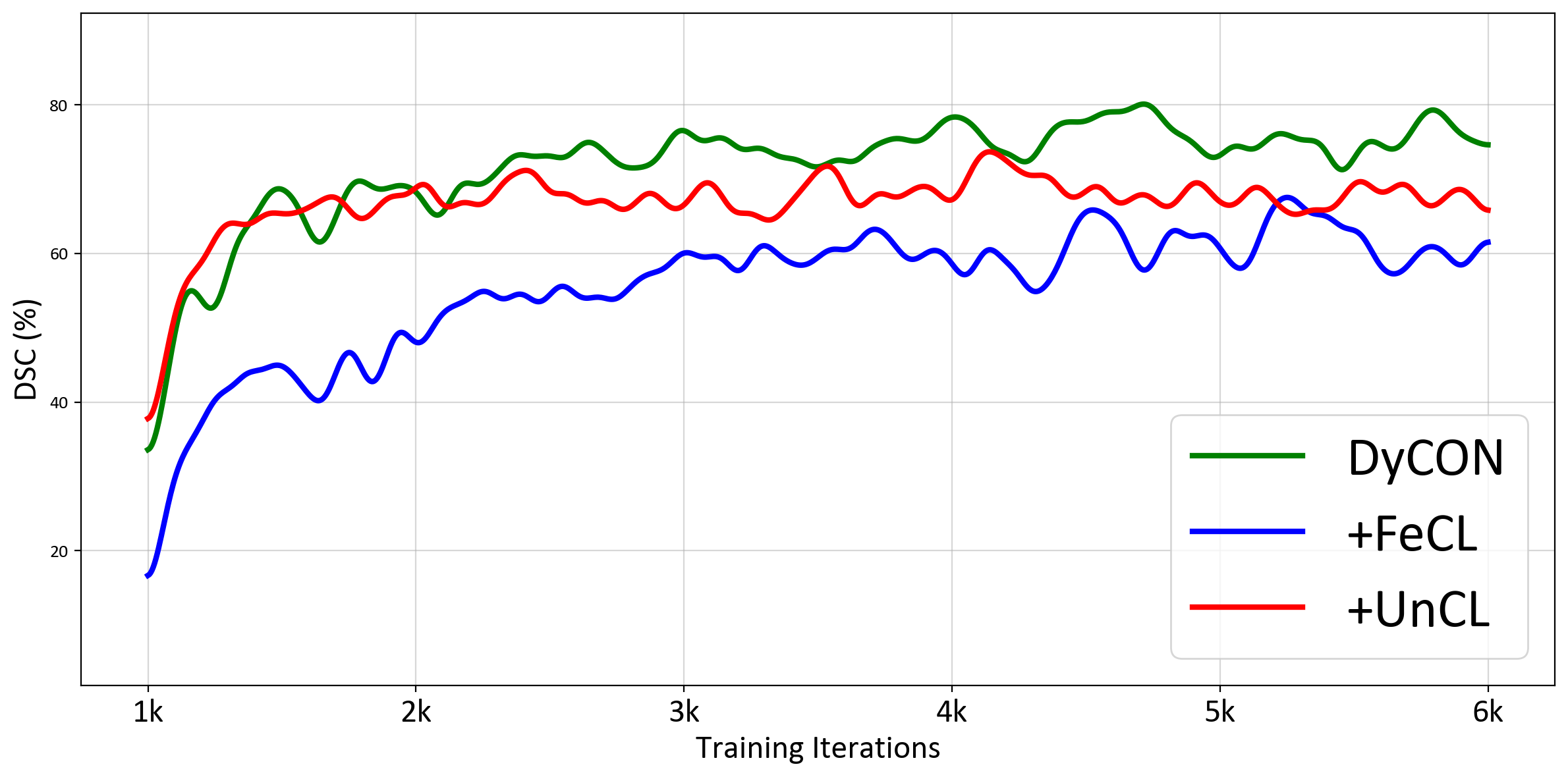} 
    \end{subfigure}
    % \hfill
    \begin{subfigure}[t]{0.46\linewidth}
        \centering
        \includegraphics[width=\linewidth]{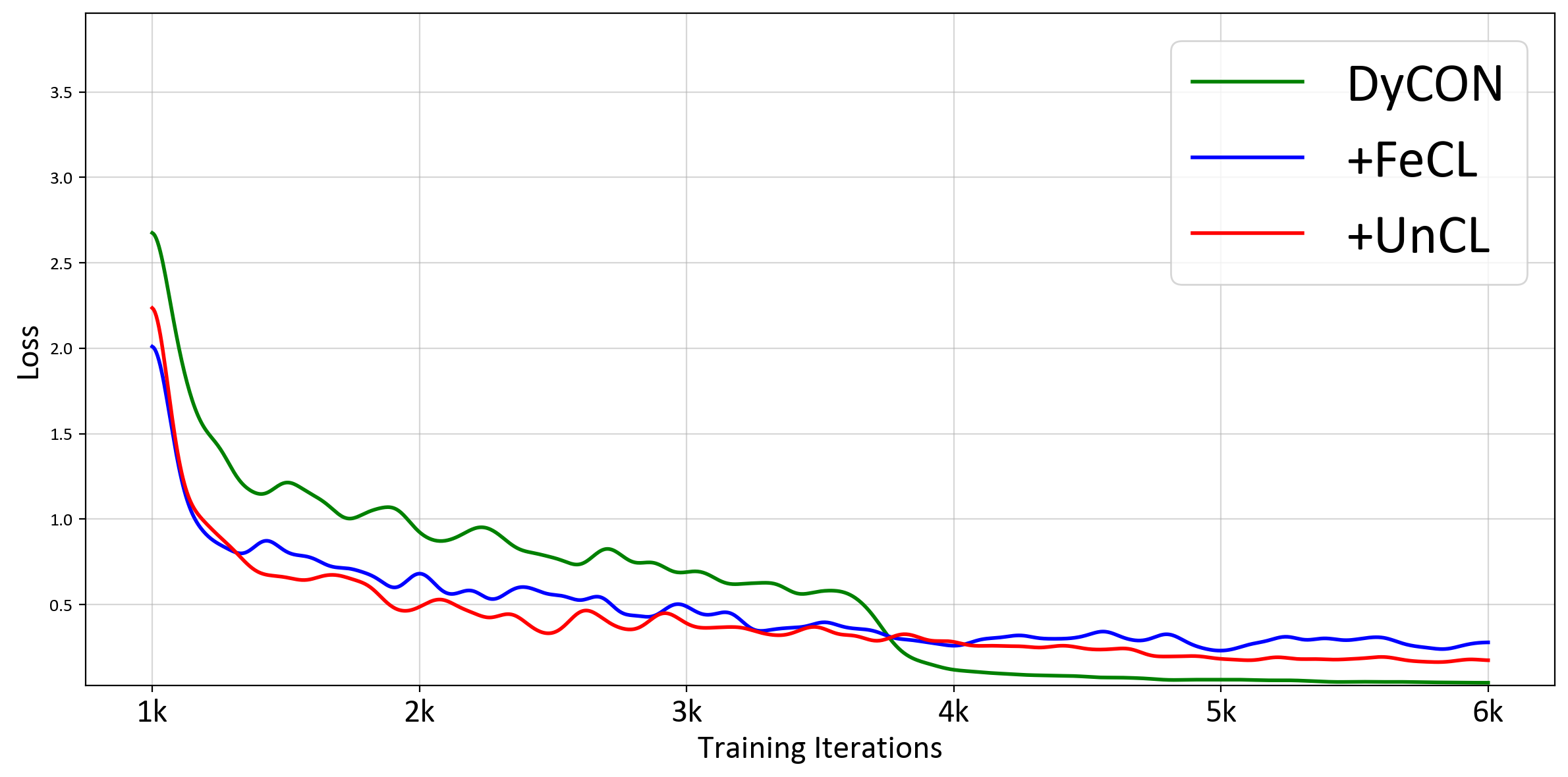} 
    \end{subfigure}
    \vspace{-0.8em}
    \caption{Loss and dice accuracy on ISLES22 with 10\% labels.}
    \vspace{-1em}
    \label{fig:loss_plot}    
\end{figure}

The results indicate that the baseline MT+SupCon model struggles with complex lesion distributions (Dice: 38.24\% on ISLES-22). Adding the focal mechanism (F$^+$+F$^-$) improves performance significantly (Dice: 63.78\%) by emphasizing hard sample pairs. Inter-network hard negatives (HN) further raise accuracy to 64.39\%, refining the model's sensitivity to subtle lesion variations. Entropy-awareness boosts Dice to 65.46\%, enhancing focus on uncertain regions. Finally, combining FeCL with UnCL yields the highest performance (Dice: 66.07\% on ISLES-22, 86.97\% on BraTS-2019), showing that integrating uncertainty-driven consistency with focused contrastive learning improves lesion delineation, handling uncertainty and class imbalance. The evolution of Dice accuracy and loss during training on the ISLES'22 dataset (with 10\% labeled data) for DyCON, DyCON+UnCL, and DyCON+FeCL is shown in Fig.~\ref{fig:loss_plot}, illustrating how each component enhances training stability and overall segmentation performance.
\begin{table}[!t]
    \centering
    \resizebox{\columnwidth}{!}{
        \begin{tabular}{cccccccccc}
        \toprule
        \multicolumn{4}{c}{FeCL Elements}   & \multicolumn{3}{c}{ISLES-2022} & \multicolumn{3}{l}{BraTS-2019} \\
        \midrule
        \multicolumn{1}{c}{F$^+$+F$^-$} & HN  & \multicolumn{1}{c}{Entropy} & \multicolumn{1}{c}{UnCL} & \multicolumn{1}{c}{Dice(\%)\(\uparrow\)} & \multicolumn{1}{c}{HD95\(\downarrow\)} & \multicolumn{1}{c}{ASD\(\downarrow\)} & Dice (\%)\(\uparrow\) & HD95\(\downarrow\) & ASD\(\downarrow\)  \\
        \midrule
        \ding{55}    & \ding{55}  & \ding{55}   & \ding{55}    & 38.24 & 20.16 & 6.35 & 82.68 & 21.53 & 5.89  \\
        \checkmark   & \ding{55}  & \ding{55}   & \checkmark   & 63.78 & 13.94 & 1.10 & 84.57 & 8.53  & 1.75 \\
        \checkmark   & \checkmark & \ding{55}   & \checkmark   & 64.39 & 13.76 & 1.00 & 85.23 & 8.11  & 1.59 \\
        \checkmark   & \ding{55}  & \checkmark  & \checkmark   & 65.46 & 13.52 & 0.85 & 86.32 & 7.86  & 1.32 \\
        \checkmark   & \checkmark & \checkmark  & \checkmark   & \textbf{66.07} & \textbf{13.34} & \textbf{0.75} & \textbf{86.97} & \textbf{7.46} & \textbf{1.16}  \\
        \bottomrule
        \end{tabular}
    }    \vspace{-0.5em}
    \caption{Ablation results on the effectiveness of FeCL components on the ISLES'22 and BraTS'19 datasets using 10\% labeled data.}
    \label{tab:FeCL_results}
    \vspace{-1em}
\end{table}

\begin{figure}[!ht]
    \centering
    \includegraphics[width=8.5cm]{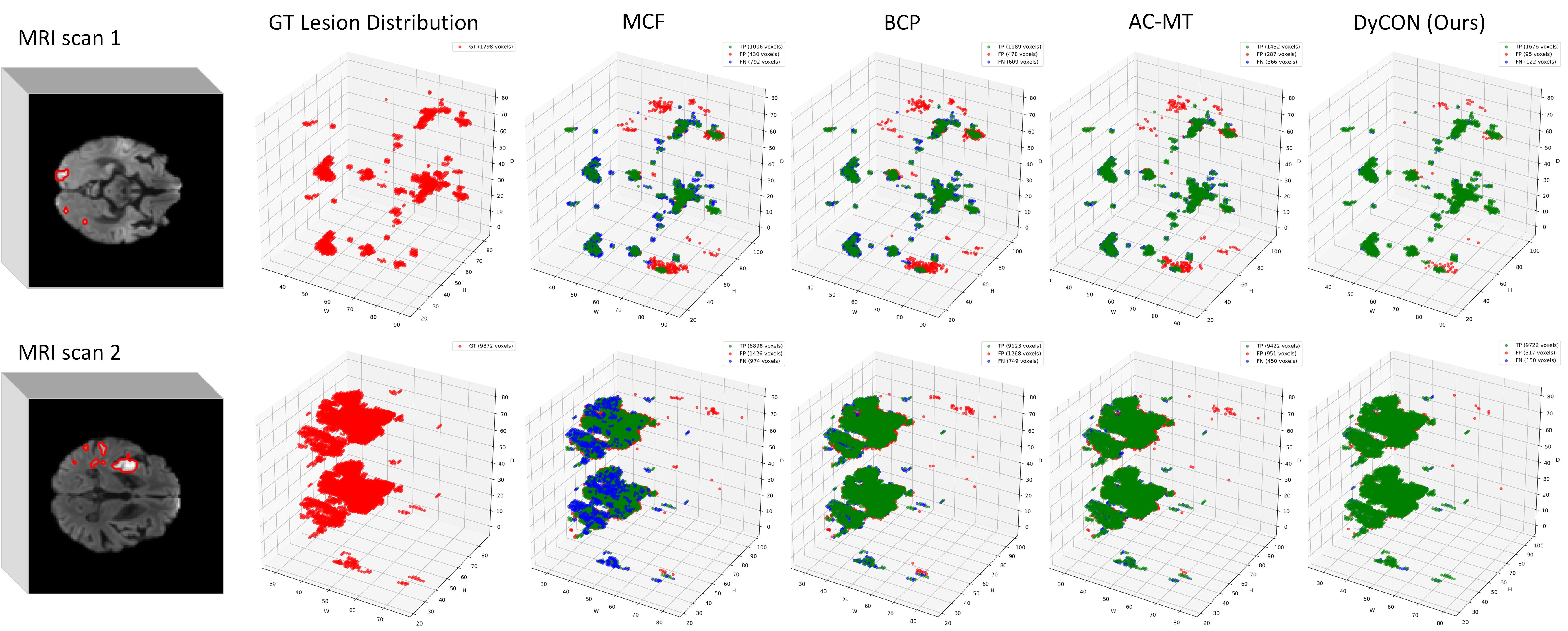}
    \caption{Visual illustration of voxel-wise lesion prediction. Existing SOTA methods struggle with detecting small and scattered lesions, resulting in higher instances of FP (red) and FN (blue) voxels while DyCON effectively reduces these errors (green).}
    \label{fig:scattered_voxels} 
    % \vspace{-1em}
\end{figure}

\textbf{Sensitivity Analysis of FeCL:} We conduct a sensitivity analysis on lesions of varying sizes and distributions in the ISLES-2022 dataset to evaluate FeCL's capacity in handling challenging brain lesion segmentation cases. The test set is categorized into \textit{small} (12 scans), \textit{medium} (3 scans), and \textit{large} (7 scans) lesions based on voxel area, as well as \textit{scattered} (73\%) and \textit{non-scattered} (27\%) lesions by spatial distribution. We compare MT models trained with SupCon and FeCL for each category using 10\% labeled data.
\begin{figure}[!t]
    \raggedright  
    \begin{subfigure}[b]{0.48\textwidth}  
        \centering
        \includegraphics[width=\textwidth]{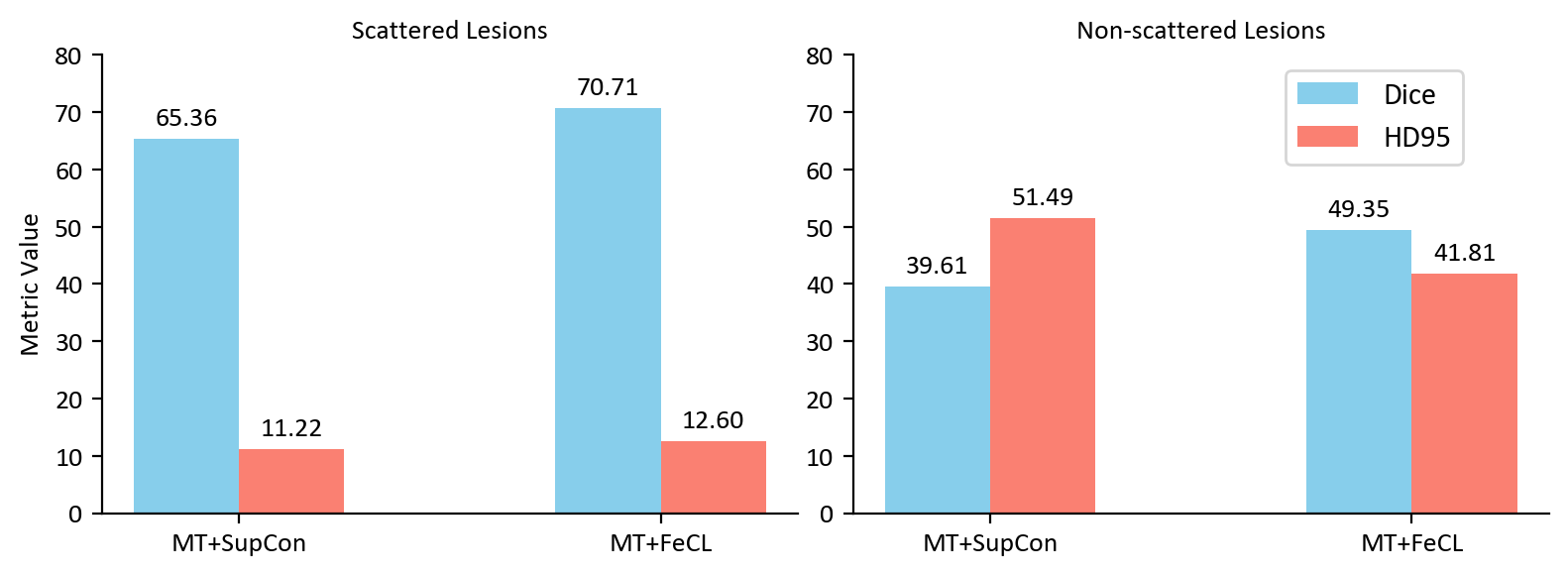} 
    \end{subfigure}
    % \vspace{0.2cm} 
    \begin{subfigure}[b]{0.48\textwidth}
        \centering
        \includegraphics[width=\textwidth]{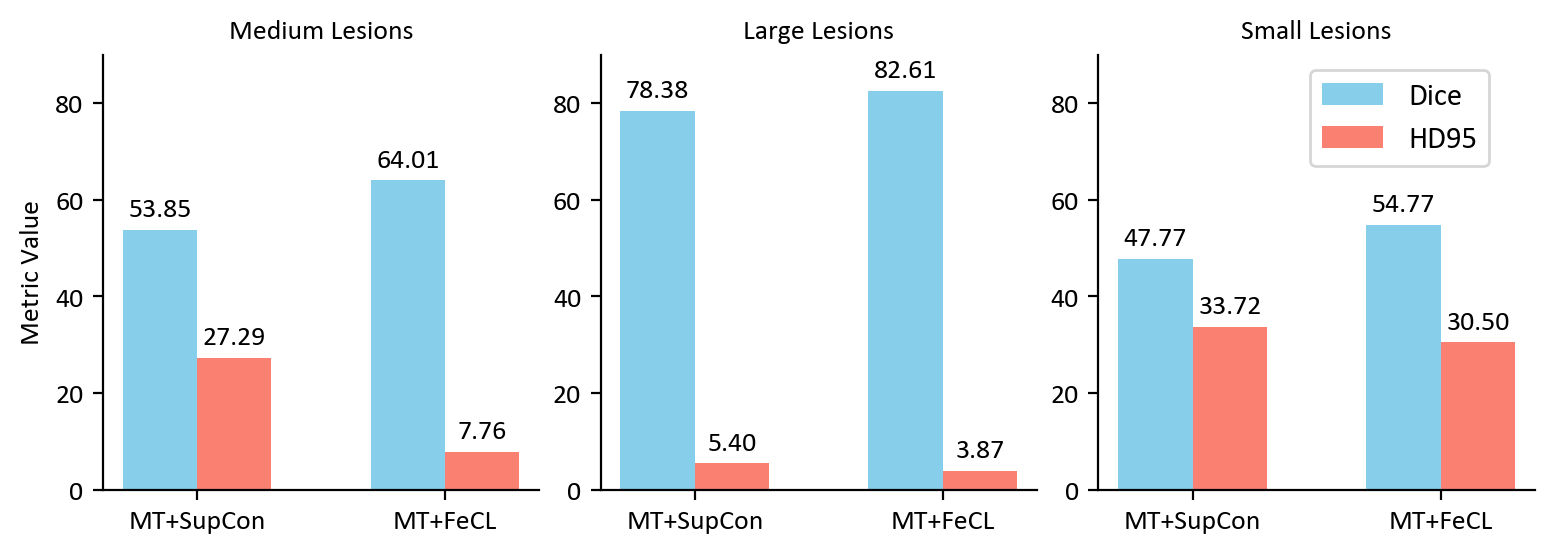}  
    \end{subfigure}
    \vspace{-1.5em}
    \caption{Comparison of Dice and HD95 metrics for SupCon and FeCL across different lesion sizes and distributions.}
    \label{fig:comparison_grouped_bar}
\end{figure}
As illustrated in Fig.~\ref{fig:comparison_grouped_bar}, FeCL consistently outperforms SupCon with small and scattered lesions, achieving higher Dice scores (54.77\% vs. 47.77\% for small, 70.71\% vs. 65.36\% for scattered) and lower HD95 and ASD values. Moreover, Fig.~\ref{fig:scattered_voxels} provides a compelling visualization of DyCON's superiority to segment scattered lesions with varying sizes. While SOTA methods struggle with false negatives and positives, DyCON remarkably identifies both large and small lesion regions, regardless of their spatial distribution.

% \paragraph{Acknowledgment} This research was funded by the Khalifa University Internal Fund and ADNOC Research Fund (Ref: 21110553).

% \vspace{-1em}
\section{Conclusion}
We presented DyCON, a semi-supervised framework to tackle uncertainty arising from lesion variations and class imbalance in medical image segmentation under limited supervision. DyCON improves the robustness of consistency frameworks with two specialized losses: UnCL that leverages entropy (a proxy for uncertainty) to guide the model to focus on ambiguous regions early in training and gradually shifting toward confident predictions; and FeCL that complements UnCL by enhancing local discrimination to emphasize hard and uncertain samples to effectively address class imbalance. Extensive experiments confirms the superiority of DyCON in boundary precision and segmentation accuracy over existing SSL methods.

\section{Acknowledgment}
% \paragraph{Acknowledgment} 
This research was funded by the Khalifa University Internal Fund and ADNOC Research Fund (Ref: 21110553).

{
    \small
    \bibliographystyle{ieeenat_fullname}
    \bibliography{main}
}

% % WARNING: do not forget to delete the supplementary pages from your submission 
% \input{sec/X_suppl}

% % \clearpage
% {
%     \small
%     \bibliographystyle{ieeenat_fullname}
%     \bibliography{main}
% }

\end{document}